\documentclass[sigconf,nonacm]{acmart}
\acmSubmissionID{304}
\usepackage{booktabs} %
\usepackage{multirow}
\usepackage{tabularx}
\usepackage{wrapfig}
\usepackage{mathtools}
\usepackage{lipsum}
\usepackage{footmisc}

\usepackage[ruled]{algorithm2e} %

\SetAlFnt{\small}
\SetAlCapFnt{\small}
\SetAlCapNameFnt{\small}
\SetAlCapHSkip{0pt}

\setcopyright{acmlicensed}

\begin{document}
\def\negativevspace{}	
\newcommand{\TODO}[1]{{\color{red}{[TODO: #1]}}}
\newcommand{\rh}[1]{{\color{green}#1}}
\newcommand{\new}[1]{{#1}}
\newcommand{\phil}[1]{{\color[rgb]{0.2,0.8,0.2}#1}}
\newcommand{\xz}[1]{{\color[rgb]{0.6,0.9,0.1}#1}}
\newcommand{\edward}[1]{{\color[rgb]{0.7,0.2,0.7}#1}}
\newcommand{\ednote}[1]{{\color[rgb]{0.7,0.2,0.7}ED:#1}}
\newcommand{\para}[1]{\vspace{.05in}\noindent\textbf{#1}}
\def\ie{\emph{i.e.}}
\def\eg{\emph{e.g.}}
\def\etal{{\em et al.}}
\def\etc{{\em etc.}}
\newcolumntype{C}[1]{>{\centering\arraybackslash}p{#1}}
	
\title{Neural Wavelet-domain Diffusion for 3D Shape Generation}

\author{Ka-Hei Hui}
\affiliation{%
	\institution{The Chinese University of Hong Kong}
	\country{HK SAR, China}}
\email{khhui@cse.cuhk.edu.hk}
\author{Ruihui Li}
\affiliation{%
	\institution{Hunan University}
	\country{China}}
\email{liruihui@hnu.edu.cn}
\author{Jingyu Hu, Chi-Wing Fu}
\affiliation{%
	\institution{The Chinese University of Hong Kong}
	\country{HK SAR, China}}
\email{{jyhu,cwfu}@cse.cuhk.edu.hk}
\renewcommand\shortauthors{Hui et al.}

\begin{teaserfigure}
\vspace*{-1mm}
  \centerline{\includegraphics[width=0.95\textwidth]{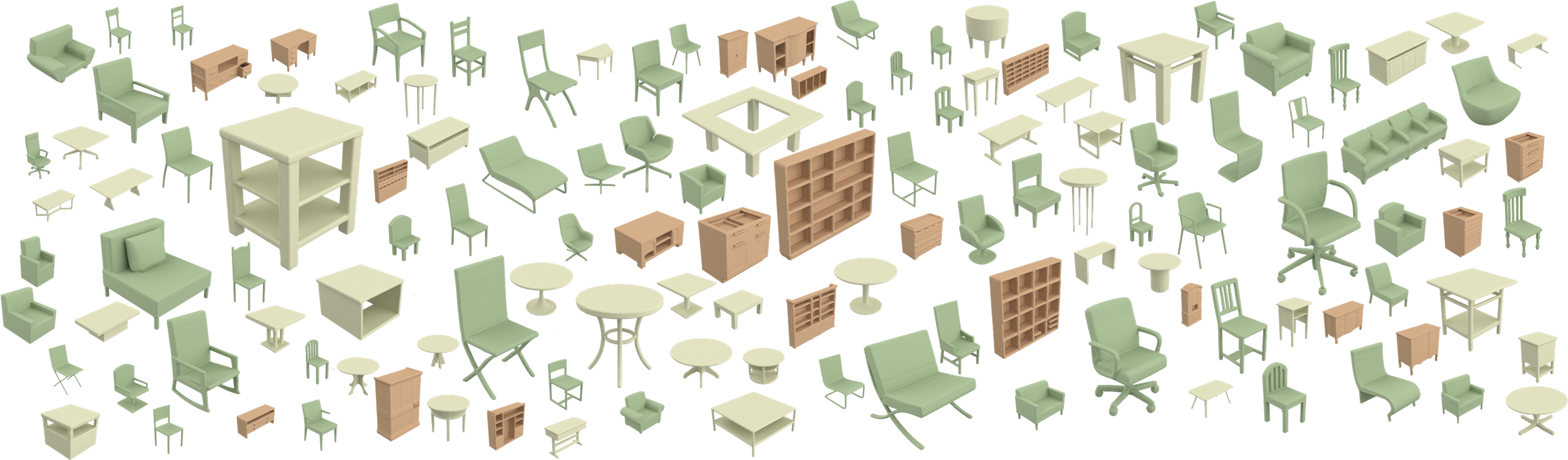}}
\vspace*{-3.5mm}
\caption{\new{Our method is able to generate diverse shapes with complex structures and topology, fine details, and clean surfaces.}
}
\label{fig:teaser}
\end{teaserfigure}

\begin{abstract}
This paper presents a new approach for 3D shape generation, enabling direct generative modeling on a continuous implicit representation in wavelet domain.
Specifically, we propose a {\em compact wavelet representation\/} with a pair of coarse and detail coefficient volumes to implicitly represent 3D shapes via truncated signed distance function\new{s} and multi-scale biorthogonal wavelet\new{s},
and formulate a pair of neural networks: 
a {\em generator} based on the diffusion model to produce diverse shapes in the form of coarse coefficient volume\new{s}; and
a {\em detail predictor\/} to further produce compatible detail coefficient volumes for enriching the generated shapes with fine \new{structures and} details.
Both quantitative and qualitative experimental results 
manifest the superiority of our approach in generating diverse and high-quality shapes with complex topology and structures, clean surfaces, and fine details, exceeding the 3D generation capabilities of the state-of-the-art models.
\end{abstract}




\maketitle

\section{Introduction}
\label{sec:intro}

Generative modeling of 3D shapes enables rapid creation of 3D contents, enriching extensive applications across graphics, vision, and VR/AR.
With the emerging large-scale 3D datasets~\cite{chang2015shapenet}, data-driven shape generation has gained increasing attention from the research community recently.
In general, a good 3D generative model should 
be able to produce diverse, realistic, 
and 
novel shapes, not necessarily the same as the existing ones.

Existing shape generation models are developed mainly for voxels~\cite{girdhar2016learning,zhu2017rethinking,yang2018learning}, point clouds~\cite{fan2017point,jiang2018gal,achlioptas2018learning}, and meshes~\cite{wang2018pixel2mesh,groueix2018papier,smith2019geometrics,tang2019skeleton}. %
Typically, these representations cannot handle
high resolutions or irregular topology, thus unlikely producing high-fidelity results.
In contrast, implicit functions~\cite{mescheder2019occupancy,park2019deepsdf,chen2019learning} show improved performance in surface reconstructions.
By representing a 3D shape as a level set of discrete volume or \new{a} continuous field, we can flexibly extract a mesh object of arbitrary topology at desired resolution.

Existing generative models such as GANs and normalizing flows have shown great success \new{in} generating point clouds and voxels.
Yet, they cannot effectively generate implicit functions.
To represent a surface in 3D, a large number of point samples are required, even though many nearby samples are \new{redundant}.
Taking the occupancy field for instance, only regions near the surface have \new{varying data values}, yet we need huge efforts to encode samples in constant and smoothly-varying regions.
Such representation non-compactness and redundancy 
demands a huge computational cost and hinders the efficiency of direct generative learning on implicit surfaces.

To address these challenges, some methods attempt to sample in a pre-trained latent space built on the reconstruction task~\cite{chen2019learning,mescheder2019occupancy} or convert the generated implicits into point clouds or voxels for adversarial learning~\cite{kleineberg2020adversarial,luo2021surfgen}. 
However, these regularizations can only be indirectly applied to the generated implicit functions, so they are not able to ensure the generation of realistic objects.
Hence, the visual quality of the generated shapes often shows a significant gap, as compared with the 3D reconstruction results, and the diversity of their generated shapes is also quite limited.

This work introduces a new approach for 3D shape generation, enabling direct generative modeling on a continuous implicit representation in the wavelet frequency domain.
Overall, we have three key contributions:
(i) a compact wavelet representation (\ie, a pair of coarse and detail coefficient volumes) based on biorthogonal wavelet\new{s} and truncated signed distance field to implicitly encode 3D shapes, facilitating effective learning of 3D shape distribution for shape generation;
(ii) a generator network formulated based on the diffusion probabilistic model~\cite{sohl2015deep} to produce coarse coefficient volumes from random noise samples, promoting the generation of diverse 
and novel
shapes; and
(iii) a detail predictor network, formulated to produce compatible detail coefficients to enhance the fine details in the generated shapes.

With the two trained networks, we can start from random noise volumes and flexibly generate diverse
and \new{realistic}
shapes \new{that are not necessarily the same as the training shapes}.
Both quantitative and qualitative experimental results manifest the 3D generation capabilities of our method, showing its superiority over \new{the}
state-of-the-art approaches.
As Figure~\ref{fig:teaser} shows, our generated shapes exhibit diverse topology, clean surfaces, sharp boundaries, and fine details, without obvious artifacts.
\new{Fine details such as curved/thin beams, small pulley, and complex cabinets 
are} very challenging for the existing 3D generation approaches \new{to synthesize}.

\vspace*{-1mm}
\section{Related Work}

\label{sec:rw}

\paragraph{3D reconstruction via implicit function.}
Recently, many methods leverage the flexibility 
of implicit surface for 3D reconstructions from 
voxels~\cite{mescheder2019occupancy,chen2019learning}, \new{complete/partial point clouds~\cite{park2019deepsdf,Liu2021MLS,yan2022shapeformer}}, and RGB images~\cite{xu2019disn,xu2020ladybird,li2021d2im,tang2021skeletonnet}.
On the other hand, besides ground-truth field values, various supervisions have been explored to train the generation of implicit surfaces,~\eg, multi-view images~\cite{liu2019learning,niemeyer2020differentiable} and unoriented point clouds~\cite{atzmon2020sal,gropp2020implicit,zhao2021sign}.
Yet, the task of 3D reconstruction focuses mainly on synthesizing a high-quality 3D shape that best matches the input.
So, it is fundamentally very different from the task of 3D shape generation, which aims to learn the shape distribution \new{of} a given set of shapes for generating diverse, high-quality, and \new{possibly novel} shapes accordingly.

\vspace*{-4pt}
\paragraph{3D shape generation via implicit function.}
Unlike 3D reconstruction, the 3D shape generation task has no fixed ground truth to supervise the generation of each shape sample.
Exploring efficient guidance for implicit surface generation is still an open problem.
Some works attempt to use the reconstruction task to first learn a latent embedding~\new{\cite{mescheder2019occupancy,chen2019learning,hao2020dualsdf,ibing20213d}} then generate new shapes by decoding 
\new{codes sampled from} the learned latent space.
Recently,~\cite{hertz2022spaghetti} learn a latent space with a Gaussian-mixture-based autodecoder for shape generation and manipulation.
Though these approaches ensure a simple training process, the generated shapes have limited diversity restricted by the pre-trained shape space.
Some other works attempt to convert implicit surfaces to some other representations,~\eg, voxels\new{~\cite{kleineberg2020adversarial,zheng2022sdfstylegan}}, point cloud~\cite{kleineberg2020adversarial}, and mesh~\cite{luo2021surfgen}, 
for applying adversarial training.
Yet, the conversion inevitably leads to information loss in the generated implicit surfaces, thus reducing the training efficiency and generation quality.

In this work, we propose to adopt a compact wavelet representation for modeling the implicit surface and learn to \new{synthesize} it with a diffusion model.
By this means, we can effectively learn to generate the implicit representation without a pre-trained latent space and a representation conversion.
The results also show that our new approach is capable of producing diversified shapes of high visual quality, exceeding the state-of-the-art methods.

\vspace*{-6pt}
\paragraph{Other representations for 3D shape generation}
\cite{smith2017improved,wu2016learning} explore voxels, a natural grid-based extension of 2D image.
Yet, the methods learn mainly coarse structures and fail to produce fine details
due to memory restriction.
Some other methods explore point clouds via GAN~\cite{gal2020mrgan,hui2020progressive,li2021spgan}, flow-based models~\cite{kim2020softflow,cai2020learning}, and diffusion models~\cite{zhou20213d}.
Due to the discrete nature of point clouds, 3D meshes reconstructed from them often contain artifacts.
This work focuses on implicit surface generation, aiming at generating high-quality and diverse meshes with fine details and
overcoming the limitations of
the existing representations.

\begin{figure*}[!t]
	\centering
	\includegraphics[width=1.0\linewidth]{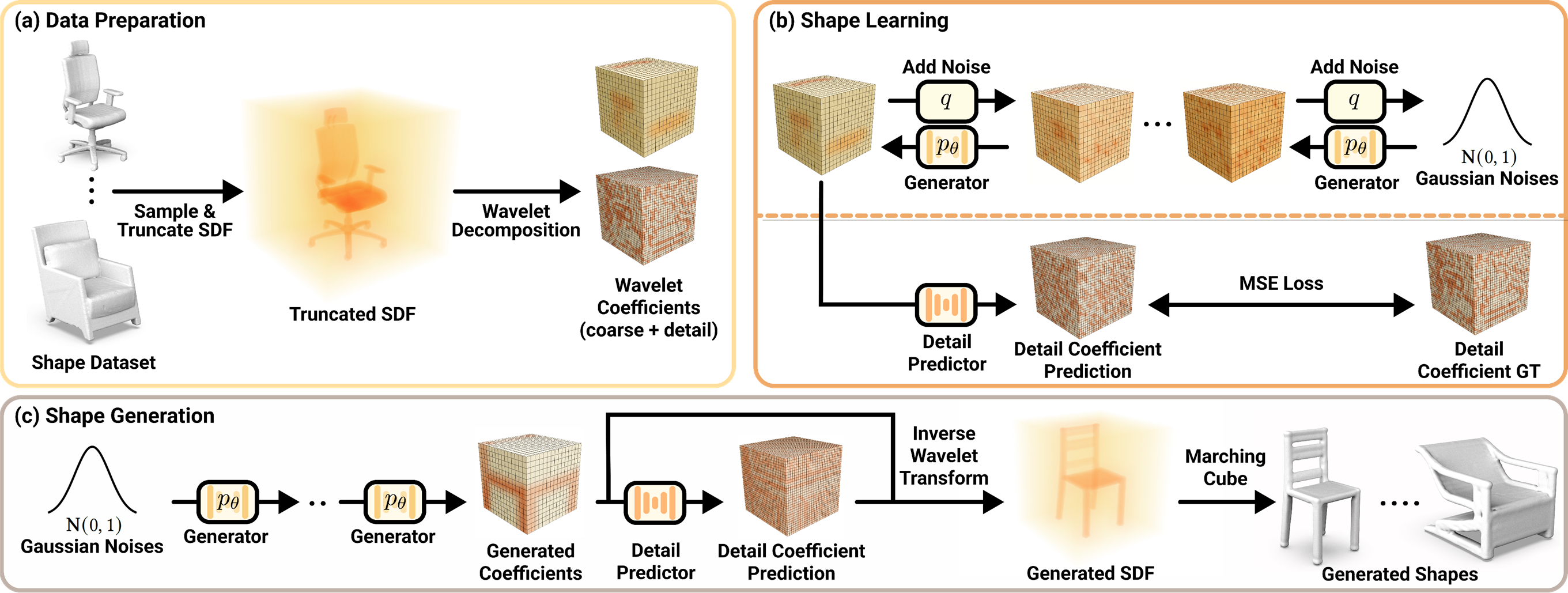}
	\vspace*{-5mm}
	\caption{Overview of our approach.
	(a) {\em Data preparation\/} builds a compact wavelet representation (a pair of coarse and detail coefficient volumes)
	for each input shape using a truncated signed distance field (TSDF) and a multi-scale wavelet decomposition.
	(b) {\em Shape learning\/} trains the generator network to produce coarse coefficient volumes from random noise samples and trains the detail predictor network to produce detail coefficient volumes from coarse coefficient volumes.
	(c) {\em Shape generation\/} employs the trained generator to produce a coarse coefficient volume and then the trained detail predictor to further predict a compatible detail coefficient volume, followed by an inverse wavelet transform and marching cube, to generate the output 3D shape.
	}
	\label{fig:overview}
	\vspace*{-1.5mm}
\end{figure*}

\vspace*{-6pt}
\paragraph{Multi-scale neural implicit representation.}
This work also relates to multi-scale representations, so we discuss some 3D deep learning works in this area.
\cite{liu2020neural,takikawa2021neural,martel2021acorn,chibane2020implicit,chen2021multiresolution} predict multi-scale latent codes in an adaptive octree
to improve the reconstruction quality and inference efficiency.
\cite{fathony2020multiplicative} propose a band-limited network
to obtain a multi-scale representation by restricting the frequency magnitude of the basis functions.
Recently,~\cite{saragadam2022miner} adopt the Laplacian pyramid to extract multi-scale coefficients for multiple neural networks.
Unlike our work, this work overfits
each input object with an individual representation for efficient storage and rendering.
In contrast to our work on shape generation,
the above methods focus on improving 3D reconstruction performance by separately handling features at different levels.
In our work, we adopt a multi-scale implicit representation based on wavelets (motivated by~\cite{velho1994multiscale}) to build a compact representation
for high-quality shape generation.

\vspace*{-6pt}
\paragraph{Denoising diffusion models.}
These models
~\cite{sohl2015deep,ho2020denoising,nichol2021improved,song2020denoising} 
recently show top performance in image generation, surpassing GAN-based models~\cite{dhariwal2021diffusion}.
Very recently,~\cite{luo2021diffusion,zhou20213d} adopt diffusion models for point cloud generation.
Yet, they fail to generate smooth surfaces and complex structures, as point clouds contain only discrete samples.
Distinctively, we adopt diffusion model with a compact wavelet representation to model a continuous signed distance field, promoting shape generation with
diverse structures and fine details.

\section{Overview}
\label{sec:overview}

Our approach consists of the following three major procedures:

\vspace*{4pt}
(i) {\em Data preparation}~is a one-time process for preparing a compact wavelet representation \new{from} each input shape; see Figure~\ref{fig:overview}(a).
For each shape, we sample a signed distance field (SDF) and 
truncate its distance values
to avoid redundant information.
Then, we transform the truncated SDF to the wavelet domain to produce a series of multi-scale coefficient volumes.
Importantly, we take {\em a pair of coarse and detail coefficient volumes\/} at the same scale as our compact wavelet representation for implicitly encoding the input shape.

\vspace*{4pt}
(ii) {\em Shape learning\/}~aims to train a pair of neural networks to learn the 3D shape distribution from the coarse and detail coefficient volumes; see Figure~\ref{fig:overview}(b).
First, we adopt the denoising diffusion probabilistic model~\cite{sohl2015deep} to formulate and train the {\em generator network\/} to learn to iteratively refine a random noise sample for generating diverse 3D shapes in the form of the coarse coefficient volume.
Second, we design and train the {\em detail predictor network\/} to learn to produce the detail coefficient volume from the coarse coefficient volume for introducing further details in our generated shapes.
Using our compact wavelet representation, it becomes feasible to train both the generator and detail predictor to successfully produce coarse coefficient volumes with plausible 3D structures and detail coefficient volumes with fine details.

\vspace*{4pt}
(iii) {\em Shape generation\/}~employs the two trained networks to generate 3D shapes; see Figure~\ref{fig:overview}(c).
Starting from a random Gaussian noise sample, we first use the trained generator to produce the coarse coefficient volume then the detail predictor to produce an associated detail coefficient volume.
After that, we can perform an inverse wavelet transform, followed by the marching cube operator~\cite{lorensen1987marching}, to generate the output 
3D shape.

\section{Method}
\label{sec:architecture}

\subsection{Compact Wavelet Representation}

\new{Preparing a compact wavelet representation from an input shape (see Figure~\ref{fig:overview}(a)) involves the following two steps}:
(i) implicitly represent the shape using a signed distance field (SDF); and
(ii) decompose the implicit representation via wavelet transform into coefficient volumes, each encoding a specific scale of the shape.

In the first step, we scale each shape to fit 
\new{$[-0.45,+0.45]^3$} and sample an SDF of resolution $256^3$ to implicitly represent the shape.
Importantly, we truncate the distance values in the SDF to 
$[-0.1,+0.1]$, so regions \new{not close to} 
object surface are clipped to a constant.
We denote the truncated signed distance field (TSDF) for the $i$-th shape in training set as $S_i$.
By using $S_i$, we can significantly reduce the shape representation redundancy and enable the shape learning process to better focus on the shape's structures and fine details.

The second step is a multi-scale wavelet decomposition~\cite{mallat1989theory,daubechies1990wavelet,velho1994multiscale} on the TSDF.
Here, we decompose $S_i$ into 
a high-frequency detail coefficient volume and 
a low-frequency coarse coefficient volume, which is roughly a compressed version of $S_i$.
We repeat this process $J$ times on the coarse coefficient volume of each scale, 
decomposing $S_i$ into a series of multi-level coefficient volumes.
We denote the coarse and detail coefficient volumes at the $j$-th step (scale) as $C^j_i$ and $D^j_i$, respectively, where $j = \{1,...,J\}$.
The representation is lossless, meaning that the extracted coefficient volumes together can faithfully
reconstruct the original TSDF via \new{a series of} inverse wavelet transforms.

\begin{figure}[!t]
	\centering
	\includegraphics[width=0.9\linewidth]{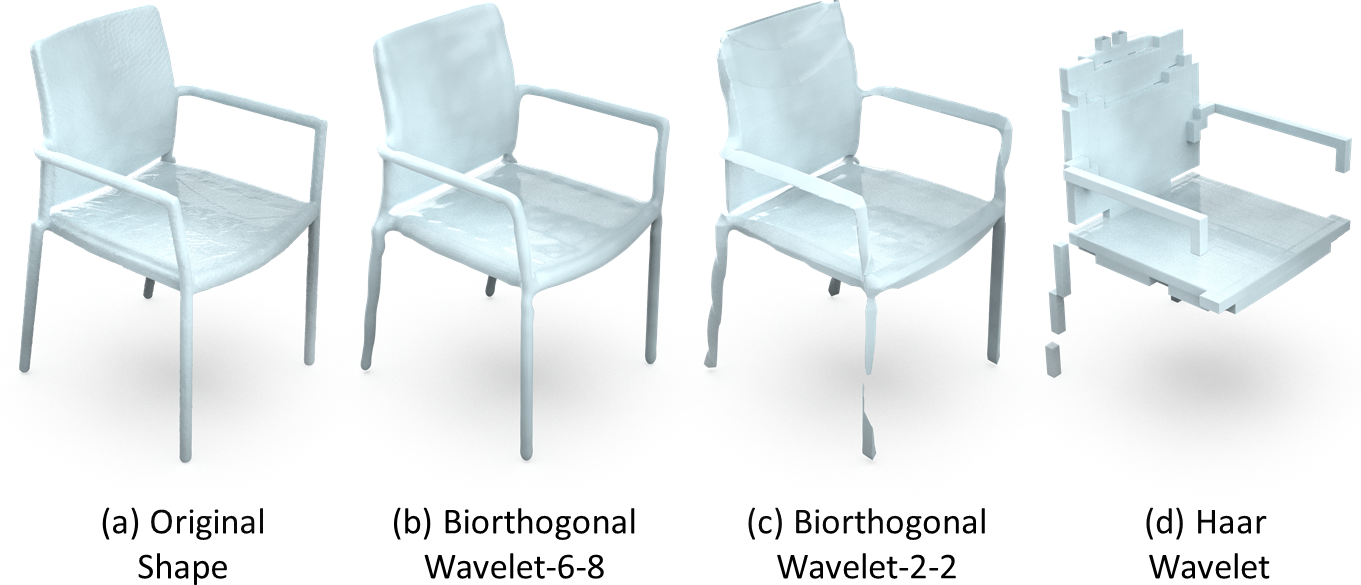}
	\vspace*{-3mm}
	\caption{Reconstructions with different wavelet filters.
	(a) An input shape from ShapeNet.
	(b,c) Reconstructions from the $J$=3 coefficient volumes with biorthogonal wavelets.
	The two numbers mean the vanishing moment of the synthesis and analysis wavelets.
	(d) Reconstruction with the Haar wavelet.}
	\label{fig:compact_analysis}
	\vspace*{-5mm}
\end{figure}

There are three important considerations in the data preparation.
First, multi-scale decomposition can effectively separate rich structures, fine details, and noise in the TSDF.
Empirically, we evaluate the reconstruction error on the TSDF by masking out all higher-scale detail coefficients and reconstructing $S_i$ only from the coefficients at scale $J=3$,~\ie, $C^3_i$ and $D^3_i$.
We found that the reconstructed TSDF values have relatively small changes from the originals (only 2.8\% in magnitude), even without 97\% of the coefficients
for the Chair category in ShapeNet~\cite{chang2015shapenet}.
Comparing Figures~\ref{fig:compact_analysis} (a) vs. (b), we can see that reconstructing only from the coarse scale of $J=3$ already well retains the chair's structure.
Motivated by this observation, we propose to construct the compact wavelet representation at a coarse scale ($J=3$)
\new{and drop other detail coefficient volumes,~\ie, $D^1_i$ and $D^2_i$,}
for efficient shape learning.
\new{More details on the wavelet decomposition are given in the supplementary material.}

Second, we need 
a suitable wavelet filter.
While Haar wavelet is a popular choice due to its simplicity,
\new{using it to encode} smooth and continuous signals such as the SDF may introduce
serious voxelization artifacts, see,~\eg, Figure~\ref{fig:compact_analysis} (d).
In this work, we propose to adopt the biorthogonal wavelet\new{s}~\cite{cohen1992biorthogonal}, since it
enables a more smooth decomposition of the TSDF.
Specifically, we tried different settings in the biorthogonal wavelet\new{s} and \new{chose} to use high vanishing moments with six for the synthesis filter and eight for the analysis filter; see 
Figures~\ref{fig:compact_analysis} (b) vs. (c).
Also, instead of storing the detail coefficient volumes with seven channels, as in traditional wavelet decomposition, we follow~\cite{velho1994multiscale} to efficiently compute it as the difference between the inverse transformed $C^{j}_i$ and $C^{j-1}_i$ in a Laplacian pyramid style.
\new{Hence, the detail coefficient volume has a higher resolution than the coarser one, but both have much lower resolution than the original TSDF volume ($256^3$).
}

Last, it is important to truncate the SDF before constructing the wavelet representation for shape learning.
By truncating the SDF, 
regions not close to the shape surface would be cast to 
a constant function to make efficient the wavelet decomposition and shape learning.
Otherwise, we found that 
the shape learning process will collapse and the training loss cannot be reduced.

\subsection{Shape Learning}
\label{ssec:shape_learning}

Next, to learn the 3D shape distribution in the given shape set, we collect coefficient volumes $\{C_i^J , D_i^J\}$ from different input shapes for training
(i) the {\em generator network\/} to learn to iteratively remove noise from a random Gaussian noise sample to generate $C_i^J$; and
(ii) the {\em detail predictor network\/} to learn to predict $D_i^J$ from $C_i^J$ to enhance the details in the generated shapes.

\vspace*{-3pt}
\paragraph{Network structure} \
To start, we formulate a simple but efficient neural network structure for both the generator and detail predictor networks.
The two networks have the same structure, since they both take a 3D volume as input and then output a 3D volume of same resolution as the input.
Specifically, we adopt a modified 3D version of the U-Net architecture~\cite{nichol2021improved}.
We first apply 3D convolution to progressively compose and downsample the input into a set of multi-scale features and a bottleneck feature volume.
Then, we apply a single self-attention layer to aggregate features in the bottleneck volume, so that we can efficiently incorporate non-local information into the features.
Further,
we upsample and concatenate features in the same scale and progressively perform an inverse convolution to produce an output of same size as the input.
Note also that for all convolution layers in the network structure, we use a filter size of three with \new{a} stride of one.

\begin{figure*}[t]
	\centering
	\includegraphics[width=0.97\linewidth]{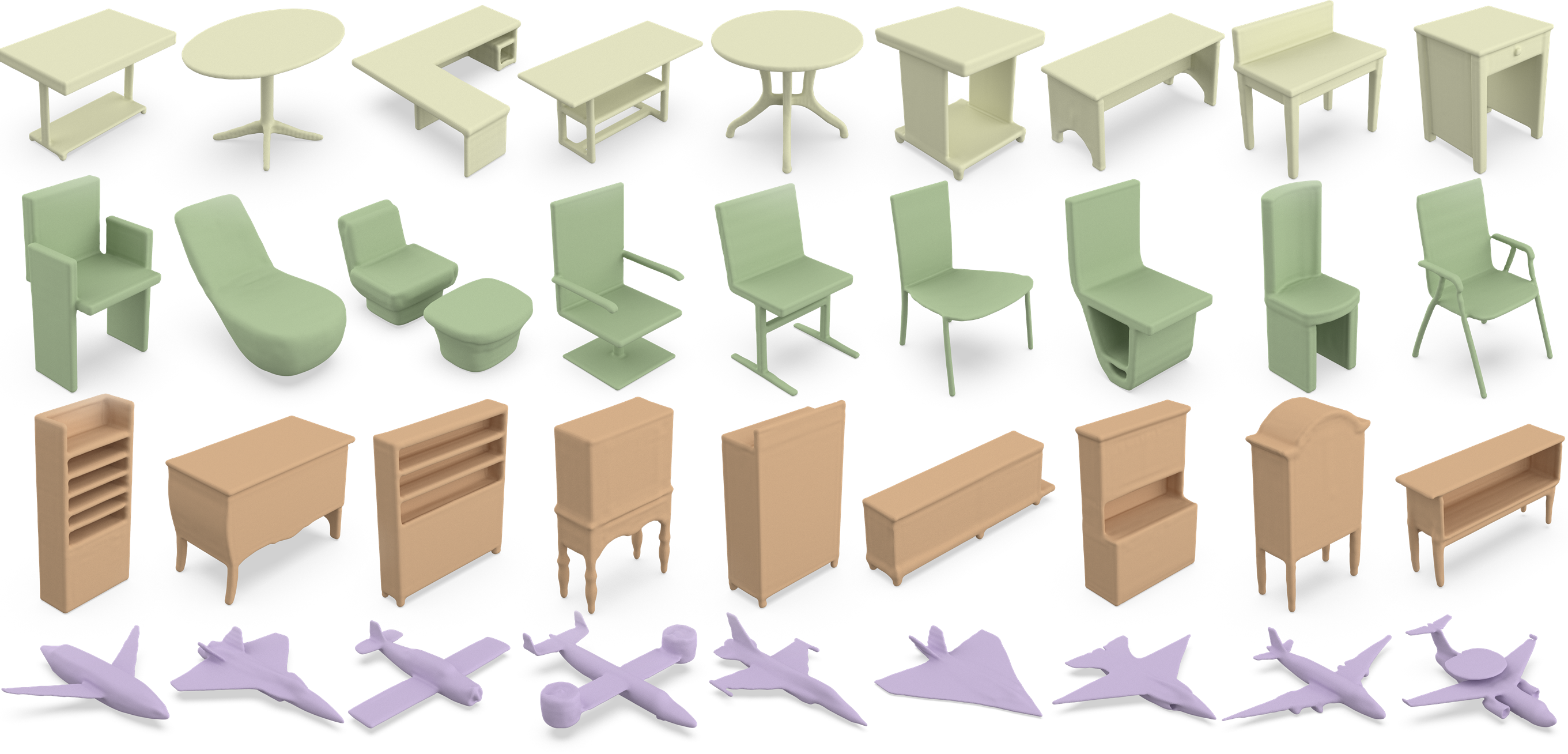}
	\vspace*{-3.5mm}
	\caption{
	Gallery of our generated shapes: Table, Chair, Cabinet, and Airplane (top to bottom).
	Our shapes exhibit complex structures, fine details, and clean surfaces, without obvious artifacts, compared with those generated by others; see Figure~\ref{fig:query_comapre}.}
	\label{fig:gallery}
	\vspace*{-3mm}
\end{figure*}

\vspace*{-3pt}
\paragraph{Modeling the generator network.} \
We formulate the 3D shape generation process based on the denoising diffusion probabilistic model~\cite{sohl2015deep}.
For simplicity, we drop the subscript and superscript in $C_i^J$ , and denote $\{ C_{0}, ..., C_{T} \}$ as the shape generation sequence, where
$C_0$ is the target, which is $C_i^J$;
$C_T$ is a random noise volume from the Gaussian prior; and
$T$ is the total number of time steps.
As shown on top of Figure~\ref{fig:overview}(b), we have 
(i) a forward process (denoted as $q(C_{0:T})$) that progressively adds noises based on a Gaussian distribution to corrupt $C_0$ into a random noise volume; and
(ii) a backward process (denoted as $p_{\theta}(C_{0:T})$) that employs the generator network (with network parameter $\theta$) to iteratively remove noise from $C_T$ to generate the target.
\new{
Note that all 3D shapes $\{ C_{0}, ..., C_{T} \}$ are represented as 3D volumes and each voxel value is a wavelet coefficient at its spatial location.
}

Both the forward and backward processes are modeled as Markov processes.
\new{
The generator network is optimized to maximize the generation probability of the target, \ie, $p_{\theta}(C_0)$.
Also, as suggested in~\cite{ho2020denoising}, this training procedure can be further simplified
to use the generator network to predict noise volume $\epsilon_{\theta}$.
Hence, we adopt 
a mean-squares loss to train our framework: 
\begin{equation}
    \label{eq:objective_simp}
    L_2 = E_{t,C_0,\epsilon}[{\parallel} \epsilon - \epsilon_{\theta}(C_t, t) {\parallel}^2], \epsilon \sim \mathcal{N}(0,\mathbf{I}),
\end{equation}
where 
$t$ is a time step;
$\epsilon$ is a noise volume; and
$\mathcal{N}(0,\mathbf{I})$ denotes a unit Gaussian distribution.
In particular, we first sample noise volume $\epsilon$ from a unit Gaussian distribution $\mathcal{N}(0,\mathbf{I})$ and a time step $t \in [1,...,T]$ to corrupt $C_0$ into $C_t$.
Then, our generator network learns to predict noise $\epsilon$ based on the corrupted coefficient volume $C_t$. 
Further, as the network takes time step $t$ as input, we convert value $t$ into an embedding via two MLP layers.
Using this embedding, we can condition all the convolution modules in the prediction and enable the generator to be more aware of the amount of noise contaminated in $C_t$.
For more details on the derivation of the training objectives, please refer to the supplementary material. 
}

% \vspace*{-3pt}

\paragraph{Detail predictor network}
With the trained generator, we can obtain diverse and good-quality coarse coefficient volumes,~\ie, $C_0$.
Next, we train the detail predictor network to produce detail coefficient volume $D_0$ from $C_0$ (see the bottom part of Figure~\ref{fig:overview}(b)), so that we can further enhance the details in our generated shapes.

To train the detail predictor network, we leverage the paired coefficient volumes $\{ C_i^J, D_i^J \}$ from the data preparation.
Importantly, detail coefficient volume $D_0$ should be highly correlated to coarse coefficient volume $C_0$.
Hence, we pose detail prediction as a conditional regression on the detail coefficient volume, aiming at learning neural network function $f: C_0 \rightarrow D_0$; hence, we optimize $f$ via
a mean squared error loss.
Overall, the detail predictor has the same network structure as the generator, but we include more convolution layers to accommodate the cubic growth in the number of 
nonzero values
in the detail coefficient volume.

\subsection{Shape Generation}
Now, we are ready to generate 3D shapes.
First, we can randomize a 3D noise volume as $C_T$ from the standard Gaussian distribution.
Then, we can employ the trained generator 
for $T$ iterations to produce $C_0$ from $C_T$.
This  process is iterative and inter-dependent.
We cannot parallelize the operations in different iterations, so leading to 
a very long computing time.
To speed up the inference process,
we adopt an approach in~\cite{song2020denoising} to sub-sample a set of time steps from $[1,..., T]$ during the inference; in practice, we evenly sample $1/10$ of the total time steps in all our experiments.

After we obtain the coarse coefficient volume $C_0$, we then use the detail predictor network to predict detail coefficient volume $D_0$ from $C_0$.
After that, we perform a series of inverse wavelet transforms from $\{ C_0 , D_0 \}$ at scale $J$=$3$ to reconstruct the original TSDF.
Hence, we can further extract an explicit 3D mesh from the reconstructed TSDF using the marching cube algorithm~\cite{lorensen1987marching}.
Figure~\ref{fig:overview}(c) illustrates the shape generation procedure.

\subsection{Implementation Details}
\label{sec:implementation}

We employed ShapeNet~\cite{chang2015shapenet} to prepare the training dataset used 
in
all our experiments.
Following the data split in~\cite{chen2019learning}, we use only the training split to supervise our network training.
Also, similar to~\cite{hertz2022spaghetti,luo2021diffusion,li2021spgan}, we train a single model for generating shapes of each category in the ShapeNet dataset~\cite{chang2015shapenet}.

We implement our networks using 
PyTorch and run all experiments on a GPU cluster with four RTX3090 GPUs.
We follow~\cite{ho2020denoising} to set $\{\beta_t\}$ to increase linearly from $1e^{-4}$ to $0.02$ for 1,000 time steps and 
set $\sigma_t = \frac{1-\bar{\alpha}_{t-1}}{1 - \bar{\alpha}_t} \beta_t$.
We train the generator for 800,000 iterations and
the detail predictor for 60,000 iterations, both 
using the Adam optimizer~\cite{kingma2014adam} with a learning rate of \new{$1e^{-4}$}.
Training the generator and detail predictor takes around three days and 12 hours, respectively.
\new{
The inference takes around six seconds per shape on an RTX 3090 GPU.
We adapt~\cite{cotter2020uses} to implement the 3D wavelet decomposition and {\em will release our code and training data upon the publication of this work.\/}
}

\begin{table*}[t]
	\centering
		\caption{Quantitative comparison between the generated shapes produced by our method and four state-of-the-art methods.
		We follow the same setting to conduct this experiment as in the state-of-the-art methods.
		From the table, we can see that our generated shapes have the best quality for almost all cases
		(lowest MMD, largest COV, and lowest 1-NNA) for both the Chair and Airplane categories.
The units of CD and EMD are $10^{-3}$ and $10^{-2}$, respectively.}
        \vspace*{-2mm}
	\resizebox{1.0\linewidth}{!}{
		\begin{tabular}{C{5cm}|C{0.6cm}C{0.6cm}C{0.6cm}C{0.6cm}C{0.6cm}C{0.6cm}|C{0.6cm}C{0.6cm}C{0.6cm}C{0.6cm}C{0.6cm}C{0.6cm}}
			\toprule[1pt]
                        \multirow{3}*{Method} & \multicolumn{6}{c|}{Chair}                                                  & \multicolumn{6}{c}{Airplane}
                        \\
                         & \multicolumn{2}{c}{COV} & \multicolumn{2}{c}{MMD} & \multicolumn{2}{c|}{1-NNA} & \multicolumn{2}{c}{COV} & \multicolumn{2}{c}{MMD} & \multicolumn{2}{c}{1-NNA} \\ 
                         & CD         & EMD        & CD         & EMD        & CD         & EMD        & CD         & EMD        & CD         & EMD        & CD         & EMD        \\ \hline
IM-GAN~\cite{chen2019learning}
& 56.49           & 54.50 & 11.79          & 14.52          & 61.98          & 63.45          & 61.55          & 62.79          & 3.320          & 8.371          & 76.21          & 76.08
\\ \hline
Voxel-GAN~\cite{kleineberg2020adversarial}
& 43.95           & 39.45 & 15.18          & 17.32          & 80.27          & 81.16          & 38.44          & 39.18          & 5.937          & 11.69          & 93.14          & 92.77
\\ \hline
Point-Diff~\cite{luo2021diffusion}
& 51.47           & \textbf{55.97} & 12.79          & 16.12          & 61.76          & 63.72          & 60.19          & 62.30          & 3.543          & 9.519          & 74.60          & 72.31
\\ \hline
SPAGHETTI~\cite{hertz2022spaghetti}
& 49.19           & 51.92 & 14.90          & 15.90          & 70.72          & 68.95          & 58.34          & 58.38          & 4.062          & 8.887          & 78.24          & 77.01          \\ \hline \hline
Ours                     & \textbf{58.19}  & 55.46 & \textbf{11.70} & \textbf{14.31} & \textbf{61.47} & \textbf{61.62} & \textbf{64.78} & \textbf{64.40} & \textbf{3.230} & \textbf{7.756} & \textbf{71.69} & \textbf{66.74} \\
			
			\bottomrule[1pt]
	\end{tabular}}
    \vspace*{-1mm}
\label{tab:quanComparison}
\end{table*}

\section{Results and Experiments}

\subsection{Galleries of our generated shapes}

Besides Figure~\ref{fig:teaser}, we present Figure~\ref{fig:gallery} to showcase the compelling 
capability of our method on generating shapes of various categories.
Our generated shapes exhibit {\em diverse topologies\/}, {\em fine details\/}, and also {\em clean surfaces without obvious artifacts\/}, covering a rich variety of small, thin, and complex structures that are typically very challenging for the existing approaches to produce.
More 3D shape generation results are provided in the supplementary material.

\vspace{-3pt}

\subsection{Comparison with Other Methods}

Next, we compare the shape generation capability of our method
with four state-of-the-art methods:
IM-GAN~\cite{chen2019learning}, 
Voxel-GAN~\cite{kleineberg2020adversarial}, 
Point-Diff~\cite{luo2021diffusion}, 
and SPAGHETTI~\cite{hertz2022spaghetti}.
To our best knowledge, ours is the first work that generates implicit shape representations in 
frequency domain and 
considers coarse and detail coefficients to enhance the generation of structures and fine details.

Our experiments follow the same setting as the above works.
Specifically, we leverage our trained model on the Chair and Airplane categories in ShapeNet~\cite{chang2015shapenet} to randomly generate 2,000 shapes for each category.
Then, we uniformly sample 2,048 points on each generated shape and evaluate the shapes using the same set of metrics as in the previous methods (details to be presented later).
As for the four state-of-the-art methods, we employ publicly-released trained network models to generate shapes.

\vspace{-3pt}
\paragraph{Evaluation metrics.} 
Following~\cite{luo2021diffusion,hertz2022spaghetti}, we evaluate the generation quality using
(i) minimum matching distance (MMD) measures the fidelity of the generated shapes;
(ii) coverage (COV) indicates how well the generated shapes cover the shapes in the given 3D repository; and
(iii) 1-NN classifier accuracy (1-NNA) measures how well a classifier differentiates the generated shapes from those in the repository.
Overall, a low MMD, a high COV, and an 1-NNA close to 50\% indicate good generation quality.
\new{More details are provided in the supplementary material.}

\begin{figure}[t]
	\centering
	\includegraphics[width=0.95\linewidth]{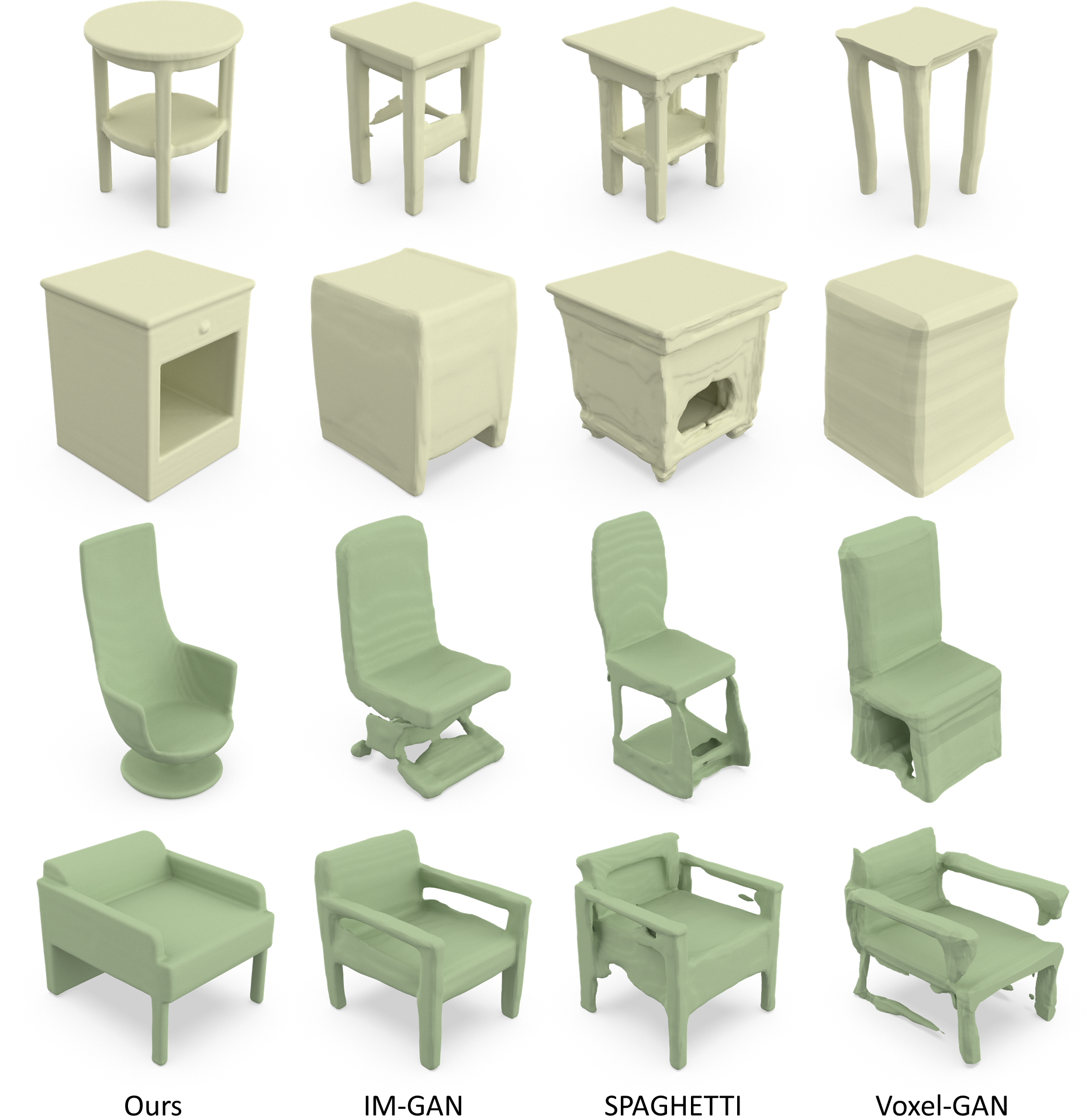}
	\vspace{-2mm}
	\caption{Visual comparisons with state-of-the-art methods.
	Our generated shapes exhibit finer details 
 and cleaner surfaces, without obvious artifacts.
	}
	\label{fig:query_comapre}
	\vspace{-2mm}
\end{figure}

\vspace{-3pt}
\paragraph{Quantitative evaluation.}

Table~\ref{tab:quanComparison} reports the quantitative comparison results,
showing that our method surpasses all others for almost all the evaluation cases over the three metrics for both the Chair and Airplane categories.
We employ the Chair category, due to its large variations in structure and topology, and the Airplane category, due to the fine details in its shapes.
As discussed in~\cite{yang2019pointflow,luo2021diffusion},
the COV and MMD metrics have limited capabilities to account for details, so they are not suitable for measuring the fine quality of the generation results,~\eg, the generated shapes sometimes show a better performance even when compared with the ground-truth training shapes on these metrics.
In contrast, 1-NNA is more robust and can better correlate with the generation quality.
In this metric, our approach outperforms all others, while having a significant margin in the Airplane category, manifesting the diversity and fidelity of our generated results.

\begin{figure}
	\includegraphics[width=0.98\linewidth]{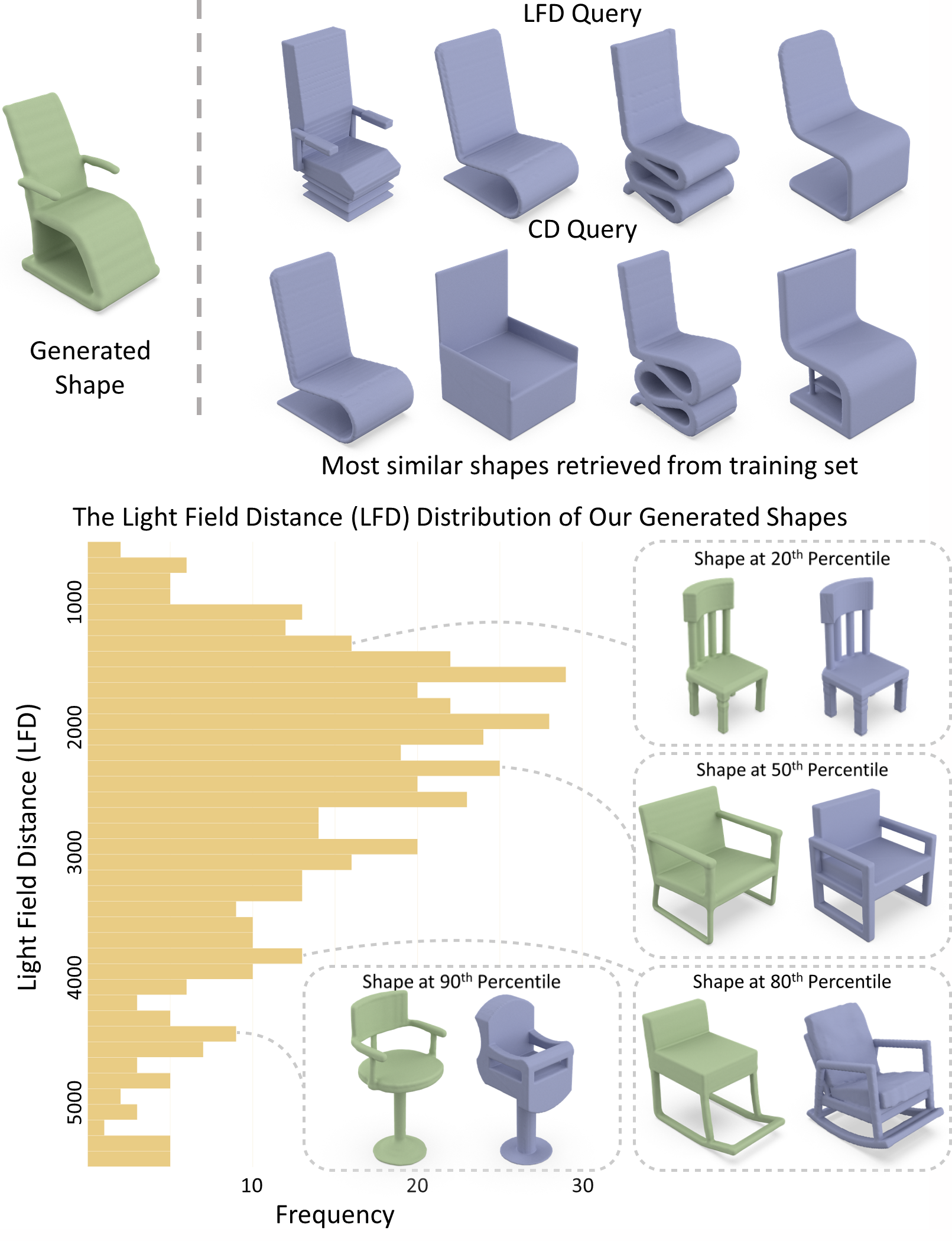}
	\vspace{-3mm}
\caption{
\new{
Shape novelty analysis.
Top: From our generated shape (in green),
we retrieve top-four most similar shapes (in blue) in training set by CD and LFD.
Bottom: We generate 500 chairs using our method;
for each chair, we retrieve the most similar shape in the training set by LFD; then, we plot the distribution of LFDs for all retrievals, showing that our method is able to generate shapes that are more similar (low LFDs) or more novel (high LFDs) compared to the training set.
Note that the generated shape at $50^{\text{th}}$ percentile is already not that similar to the associated training-set shape.
}
	\vspace{-1mm}
}
	\label{fig:novelty_analysis}
\end{figure}

\vspace{-3pt}
\paragraph{Qualitative Evaluation}
Figure~\ref{fig:query_comapre} show some visual comparisons.
For each random shape generated by our method, we find a similar shape (with similar structures and topology) generated by each of the other methods to make the visual comparison easier.
\new{See supplementary material Sections B and D for more visual comparisons.}
\new{Further, as different methods likely
have different statistical modes in the shape generation distribution, we also 
take random shapes generated by IM-GAN and find similar shapes generated by our method for comparison; see supplementary material Section~C for the results.
}
From \new{all} these results, we can see that the 3D shapes generated by our method clearly exhibit finer details, higher fidelity structures, and cleaner surfaces, without obvious artifacts.

\subsection{Model Analysis}

\paragraph{Shape novelty analysis.}
\new{Next, we analyze whether our method can generate shapes that are not necessarily the same as the training-set shapes, meaning that it does not simply memorize the training data.}
To do so, we use our method to generate 500 random shapes
and retrieve top-four most similar shapes in the training set separately via two different metrics,~\ie, Chamfer Distance (CD) and \new{Light Field Distance (LFD)~\cite{chen2003visual}.
\new{It is noted that LFD is computed based on rendered images from multiple views on each shape, so it focuses more on the visual similarity between shapes and is considered to be more robust for shape retrieval.
For the details on the metrics, please see the supplementary material.
}
}

\new{
Figure~\ref{fig:novelty_analysis} (top) shows a shape generated by our method, together with top-four most similar shapes retrieved from the training set by CD and LFD; due to the page limit, another ten
examples are shown in the supplementary material.
}
Comparing our shapes with the retrieved ones,
we can see that the shapes share similar structures, showing that our method is able to generate realistic-looking structures like those in the training set.
Beyond that, our shapes exhibit noticeable differences in various local structures.

\new{
As mentioned earlier, a good generator should produce diverse shapes that are not necessarily the same as the training shapes. 
So, we further statistically analyze the novelty of our generated shapes relative to the training set.
To do so, we use our method to generate 500 random chairs; for each generated chair shape, we use LFD to retrieve the most similar shape in the training set.
Figure~\ref{fig:novelty_analysis} (bottom) plots the distribution of LFDs between our generated shapes (in green) and retrieved shapes (in blue).
Also, we show four shape pairs at various percentiles, revealing that shapes with larger LFDs are more different from the most similar shapes in the training set.
From the LFD distribution, we can see that our method can learn a generation distribution that covers shapes in the training set (low LFD) and also generates novel and realistic-looking shapes that are more different (high LFD) from the training-set shapes.
}

\vspace*{-4pt}
\paragraph{Ablation Study}
\new{To evaluate the major components in our method, we conducted an ablation study by successively changing our full pipeline.
First, we evaluate the generation performance with/without the detail predictor.
Next, we study the importance of the diffusion model and the wavelet representation in the generator network.

\begin{table}[t]
	\centering
		\caption{Comparing our full pipeline with various ablated cases on the Chair category.
		The units of CD and EMD are $10^{-3}$ and $10^{-2}$, respectively.
  }	%

		\vspace*{-2mm}
		\resizebox{0.95\linewidth}{!}{
		\begin{tabular}{C{3cm}|C{0.5cm}C{0.5cm}C{0.5cm}C{0.5cm}C{0.5cm}C{0.5cm}}
			\toprule[1pt]
			\multirow{2}{*}{Method} 
			& \multicolumn{2}{c}{COV $\uparrow$}
			& \multicolumn{2}{c}{MMD $\downarrow$}
			& \multicolumn{2}{c}{1-NNA $\downarrow$} \\
			& CD & EMD & CD & EMD & CD & EMD \\ \hline
			Full Model & \textbf{58.19} & \textbf{55.46} & \textbf{11.70} & \textbf{14.31} & \textbf{61.47} & \textbf{61.62} \\ \hline
			W/o detail predictor & 54.20 & 50.96 & 12.32 & 14.54 & 62.46 & 62.57 \\
			VAD Generator & 21.83 & 26.77 & 21.83 & 26.77 & 95.20 & 93.62 \\
			 Direct predict TSDF & 50.51 & 50.67 & 12.83 &15.24 & 68.69 & 68.29 \\
			\bottomrule[1pt]
	\end{tabular}}
	\label{tab:analysis}
	\vspace{-2mm}
\end{table} 

The results in Table 2 demonstrate the capability of the detail predictor, which introduces a substantial improvement on all metrics (first vs. second rows).
Further, replacing our generator with the VAD model or directly predicting TSDF leads to a performance degrade (second \& last two rows).
Due to the page limit, please refer to the supplementary material for the details on how the ablation cases are implemented and the visual comparison results.

\vspace*{-4pt}
\paragraph{Limitations.}
Due to the page limit, please refer to Section~K of the supplementary material for the discussion on limitations.
}

\section{Conclusion}

This paper presents a new generative approach for learning 3D shape distribution and generating diverse, high-quality, and \new{possibly novel} 3D shapes.
Unlike prior works, we operate on the frequency domain.
By decomposing the implicit function in the form of TSDF using biorthogonal wavelet\new{s}, 
we build a compact wavelet representation with a pair of coarse and detail coefficient volumes, as an encoding of 3D shape.
Then, we formulate our generator upon
a probabilistic diffusion model to learn to generate diverse shapes in the form of coarse coefficient volumes from noise samples, and a detail predictor to further learn to generate compatible detail coefficient volumes for reconstructing fine details.
Both quantitative and qualitative experimental results demonstrate the superiority of our method in generating diverse
and realistic shapes that exhibit fine details, complex and thin structures, and clean surfaces, beyond the generation capability of the state-of-the-art methods.

To our best knowledge, this is the first work that successfully adopts a compact wavelet representation for an unconditional generative modeling on 3D shape generation, enabling many directions for future research.
At first glance, our benefits can be extended to other downstream tasks with extra conditions,~\eg, shape reconstruction from images or point clouds, and shape editing with user inputs.
Another promising future direction is to adopt wavelet-based 3D generation to animation production,~\eg, generating sequences of character motion with spatio-temporal wavelet representations.
Also, we would like to explore 
more challenging cases,~\eg, objects with extremely fine details and generation of 3D scenes.

\begin{acks}
\new{The authors would like to thank the anonymous reviewers for their valuable comments.
We also acknowledge help from Tianyu Wang for various visualizations in the paper.
This work is supported by Research Grants Council of the Hong Kong Special Administrative Region (Project No. CUHK 14206320 \& 14201921) and National Natural Science Foundation of China (No. 62202151).
} 
\end{acks}

\bibliographystyle{ACM-Reference-Format}
\bibliography{bibliography}

\end{document}